\documentclass[10pt,twocolumn,letterpaper]{article}

\usepackage[pagenumbers]{cvpr} 
\usepackage{amsmath}
\allowdisplaybreaks
\usepackage{microtype} 
\definecolor{cvprblue}{rgb}{0.21,0.49,0.74}
\usepackage[pagebackref,breaklinks,colorlinks,allcolors=cvprblue]{hyperref}
\usepackage{circledsteps}
\usepackage{colortbl}
\usepackage{multirow}
\usepackage{booktabs}
\usepackage{pifont}%
\usepackage[most]{tcolorbox}
\tcbset{
  colback=gray!15,
  colframe=gray!70,
  boxrule=0pt,
  arc=2mm,
  breakable=true
}

\definecolor{green}{rgb}{0,0.6,0}
\definecolor{red}{rgb}{0.8, 0.25, 0.33}

\newcommand{\task}{\textsc{Spotlight}}

\usepackage[most]{tcolorbox} 
\newenvironment{prompt}
  {\begin{tcolorbox}[colback=gray!4,colframe=gray!35,boxrule=0.4pt,
                     left=6pt,right=6pt,top=6pt,bottom=6pt]}
  {\end{tcolorbox}}
\usepackage[linesnumbered,ruled,vlined]{algorithm2e}

\def\paperID{22419} 
\def\confName{CVPR}
\def\confYear{2026}

\title{\task: Identifying and Localizing Video Generation Errors Using VLMs}

\author{Aditya Chinchure$^{1,2}$  \hspace{5pt}
   Sahithya Ravi$^{1,2}$ \hspace{5pt}
  Pushkar Shukla$^{3}$ \\
  Vered Shwartz$^{1,2}$ \hspace{5pt}
  Leonid Sigal$^{1,2}$\\
   [1em]
    $^1$The University of British Columbia, \hspace{5pt} 
    $^2$Vector Institute of AI \hspace{5pt} \\
    $^3$The Wharton School, University of Pennsylvania  \\
    \texttt{\small \{aditya10, sahiravi\}@cs.ubc.ca} 
}

\begin{document}

\twocolumn[{
    \renewcommand\twocolumn[1][]{#1}
    \vspace{-1em}
    \maketitle
    \vspace{-1em}
    \begin{center}
        \centering
        \includegraphics[width=1.0\textwidth]{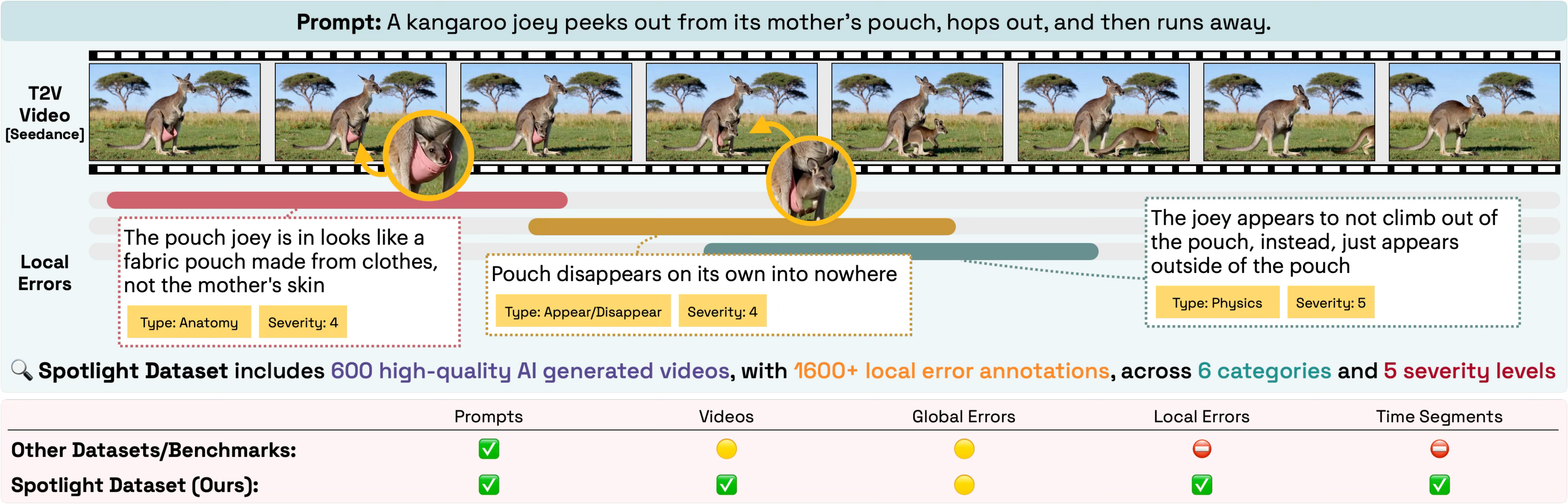}
        \captionof{figure} {We introduce \task, a benchmark for testing vision-language models on identifying and localizing errors in AI-generated videos. To the best of our knowledge, it is the first dataset with fine-grained, temporally localized and global error annotations. The dataset includes 1,604 localized error instances across 600 videos and six error categories, providing a detailed view of subtle inconsistencies in otherwise realistic generations.}
        \label{fig:introteaser}
    \end{center}
}]

\begin{abstract}

Current text-to-video models (T2V) can generate high-quality, temporally coherent, and visually realistic videos. 
Nonetheless, errors still often occur, and are more nuanced and local compared to the previous generation of T2V models. While current evaluation paradigms assess video models across diverse dimensions, they typically evaluate videos holistically without identifying when specific errors occur or describing their nature. We address this gap by introducing \task, a novel task aimed at localizing and explaining video-generation errors. We generate 600 videos using 200 diverse textual prompts and three state-of-the-art video generators (Veo 3, Seedance, and LTX-2), and annotate over 1600 fine-grained errors across six types, including motion, physics, and prompt adherence. We observe that adherence and physics errors are predominant and persist across longer segments, whereas appearance-disappearance and body pose errors manifest in shorter segments. We then evaluate current VLMs on \task\ and find that VLMs lag significantly behind humans in error identification and localization in videos. We propose inference-time strategies to probe the limits of current VLMs on our task, improving performance by nearly 2$\times$. Our task paves a way forward to building fine-grained evaluation tools and more sophisticated reward models for video generators.
\end{abstract}

\section{Introduction}
\label{sec:intro}

Video generation has rapidly advanced in recent years, driven by large-scale diffusion and transformer-based architectures. Modern text-to-video (T2V) models such as Veo 3 \cite{veo2024}, Seedance \cite{Gao2025Seedance1E}, and LTX-2 \cite{hacohen2024ltxvideorealtimevideolatent} can now produce high-quality, temporally coherent, and visually realistic outputs.  However, as these models have overcome the globally noticeable artifacts of earlier systems \cite{huang2023vbench, wang2024storyeval}, they now exhibit a more subtle class of errors that are localized and difficult to detect.

These local errors show momentary inconsistencies in videos, ranging from morphing objects and deforming body parts, to interactions that break physical plausibility. Consider the example in Fig. \ref{fig:introteaser}, where a baby kangaroo hops out of its mother's pouch. Although this video generated by Seedance looks aesthetically pleasing, it shows the mother's pouch as being made of \textit{fabric}, and eventually \textit{disappearing} as the baby kangaroo runs away. Such inconsistencies are challenging to detect because they are transient, context-dependent, and seem visually coherent. Despite their subtlety, they can disrupt human immersion, often forcing viewers to instinctively rewatch and scrutinize the problematic moment just to understand what went wrong.



%
Recent work has focused on evaluating video generation quality across dimensions such as realism, temporal consistency, and semantic alignment \cite{zheng2025vbench2, wang2024storyeval}, standardizing comparisons and driving progress in diffusion-based video models. However, most benchmarks overlook localized errors, and instead consider global metrics that capture overall appearance but fail to reflect \textit{how} and \textit{where} errors occur within a clip. Without explicit localization, errors may be left undetected, despite the video containing brief distortions, object swaps, or implausible interactions that disrupt coherence and immersion. Vision-language models (VLMs) with video understanding capabilities offer a strong foundation for building metrics with error localization, but no prior works \cite{zhang2025videorewardbench, liu2025videoalignimproving, han2025videobench} have evaluated their capability at this task.

To address this gap, we introduce \task, the first dataset and benchmark for localizing and explaining fine-grained errors in high-quality videos generated by T2V models. \task\ focuses on identifying \textit{when} and \textit{why} T2V models fail, providing interpretable and local explanations for errors in videos. We collect 1,604 human-labeled errors over a dataset of 600 videos from 200 diverse prompts, including both creative and realistic content, generated using three state-of-the-art T2V models. Our error annotations span six categories, including physics violations, appearance inconsistencies, and adherence. Each error includes temporal localization, marked by start and end times, and a severity score capturing its visual and semantic impact. Our dataset reveals that localized errors are common across all three T2V models, even when global quality is high.

We benchmark open and closed-source VLMs on \task\ and find that while current models can detect broad anomalies, they struggle with fine-grained temporal reasoning and precise localization. To address these limitations, we explore inference-time decomposition strategies using Qwen3-VL and show that a simple multi-agent approach doubles its base performance, matching significantly larger models like Gemini 2.5 Pro. Yet, our experiments reveal that humans consistently outperform all models by 2$\times$, in both precise localization and error reasoning, indicating that current VLMs struggle with the visual perception, reasoning and localization abilities required for \task. 

\task\ enables new insights into T2V failures through our dataset, and provides a novel and challenging benchmark for VLMs on fine-grained video understanding and localization with out-of-distribution AI-generated videos. VLMs capable at \task\ can not only be effective T2V evaluation models, but can also be used for curating and filtering T2V generated content, as reward models with fine-grained error attribution for training video generators, and for developing real-versus-fake video detectors for safety. \task\ provides the necessary foundation for developing automated video evaluators that align closely with human perception, and work across any T2V model and prompt.

\section{Related Work}

\subsection{Text-to-Video Models}
Recent progress in text-to-video (T2V) generation has been driven by advances in large-scale diffusion and transformer-based architectures. The prior generation of T2V models, such as LTX Video \cite{hacohen2024ltxvideorealtimevideolatent} and CogVideoX \cite{yang2025cogvideoxtexttovideodiffusionmodels}, demonstrated significant capabilities but produced videos with notable limitations in quality and generation fidelity.  However, modern T2V models have achieved substantial improvements. Commercial systems like Google's Veo 3 \cite{veo2024} can now generate high-quality videos with improved prompt adherence and realistic physics. ByteDance's Seedance 1.0 \cite{Gao2025Seedance1E} demonstrates excellent capabilities in handling complex scenarios, multi-shot generation, and long-range temporal coherence. Despite their success in generating visually appealing content, such models still exhibit errors in maintaining object identity, realistic motion, and physical plausibility, especially in complex or long-duration scenes \cite{wang2024storyeval, bansal2024videophy}. Our work investigates these errors by detecting, describing, and temporally localizing them within generated videos.

\subsection{Video Generation Evaluation}

Evaluating video generation quality requires multiple dimensions of assessment. Early benchmarks used metrics like FVD \cite{unterthiner2019fvd}, IS \cite{wu2024bettermetrictexttovideogeneration}, and CLIP-based similarity \cite{radford2021learningtransferablevisualmodels} to measure fidelity, text-video alignment, and motion smoothness. EvalCrafter \cite{liu2024evalcrafter} expanded metrics to include visual question answering and flow-based methods.
VBench \cite{huang2023vbench} and VBench 2.0 \cite{zheng2025vbench2} expanded evaluation to 18 dimensions, including physics consistency and controllability, to benchmark and rank T2V models. Specialized benchmarks like Physics-IQ \cite{motamed2025generative} and VideoPhy2 \cite{bansal2024videophy} assess physical reasoning, and StoryEval \cite{wang2024storyeval} measures prompt adherence for complex scenes. Many of these works use VLMs to evaluate videos, but only provide global feedback. 

Another line of works have introduced benchmarks and methods for building VLM-based reward models, including VideoRewardBench \cite{zhang2025videorewardbench}, MJ-Video \cite{chen2024mj}, VideoAlign \cite{liu2025videoalignimproving} and Video-Bench \cite{han2025videobench}. Our work complements these approaches, with the added focus on temporal localization. Recent datasets attempt finer-grained annotation, but address different problems from ours. BrokenVideos \cite{lin2025brokenvideosbenchmarkdatasetfinegrained} annotates videos for low-level artifacts like distortions and implausible object trajectories. DAVID-XR1 \cite{gao2025davidxr1detectingaigeneratedvideos} provides frame-level annotations with rationales but focuses on detecting AI-generated content rather than understanding generation failures.
\task\ addresses these gaps by providing temporal localization and natural language explanations of high-level semantic errors that characterize modern high-quality generators.

\section{The \task\  Task}
\label{sec:task}


We introduce \task, a task designed to identify errors that violate realism principles while respecting the creative constraints established by the prompt. For a given AI-generated video $V$ produced from prompt $P$ with duration $d$ seconds, the \task\ task seeks to identify errors that violate realism principles. We define an error as a segment of the video where the content exhibits logical inconsistencies, physical implausibilities, anatomical defects, or motion artifacts that are not justified by the prompt's creative premise. 
We categorize errors into six distinct types, as detailed in Table~\ref{tab:errors}. These categories encompass the most common failure modes observed in AI-generated videos. 

The goal of \task\ is to return a set of errors $E = \{e_1,... e_i,...\}$ such that each error $e_i = (t^s_i, t^e_i, c_i, r_i)$ is composed of the start and end times $t^s_i$ and $t^e_i$, the label $c_i$ (Table~\ref{tab:errors}) and a natural language reasoning for the error, $r_i$. Two interdependent components are necessary:


\noindent\textbf{Error Localization.} We want to determine the temporal boundaries $(t^s_i, t^e_i)$ where each error $e_i$ occurs in the video. Since errors are often localized to specific temporal segments rather than persisting throughout the entire video, precise temporal localization is essential. This component identifies \textit{where} in the temporal dimension the error manifests, enabling targeted analysis. Multiple errors may have overlapping or disjoint temporal segments, and a single frame may contain multiple distinct errors.
    
\noindent \textbf{Error Reasoning.} For each localized segment $(t^s_i, t^e_i)$, we want to determine its type $c_i$ and generate the natural language reason $r_i$. This reason must explain \textit{what} the error is and \textit{why} it is a genuine generation failure rather than an intentional creative choice justified by the prompt $P$. This $r_i$ must be grounded in the content of the video.

\begin{table}[t]
\centering
\small
\setlength{\tabcolsep}{4pt}

\begin{tabular}{@{}p{0.22\columnwidth}p{0.70\columnwidth}@{}}
\toprule
\textbf{Error Type} & \textbf{Description} \\ 
\midrule
\textbf{Physics} & Violations of physical laws, e.g., floating objects or unnatural trajectories. \\
\addlinespace[2pt]
\textbf{App/Disapp} & Objects, people, or backgrounds appear or disappear unnaturally between frames. \\
\addlinespace[2pt]
\textbf{Logical} & Actions that cannot logically co-exist, e.g., a person holding two poses simultaneously. \\
\addlinespace[2pt]
\textbf{Motion} & Unnatural motion such as objects passing through each other or jittery movement. \\
\addlinespace[2pt]
\textbf{Anatomy} & Impossible body shapes or joint movements; unrealistic morphing. \\
\addlinespace[2pt]
\textbf{Adherence} & Failure to follow key elements or intent of the text prompt. \\
\bottomrule
\end{tabular}
\caption{Taxonomy of error types in AI-generated videos.}
\label{tab:errors}
\end{table}

\section{The \task\ Dataset}
\label{sec:dataset}
\subsection{Data Collection}
To create a dataset for \task\, we first aggregated a set of diverse prompts, generated videos using these prompts, and collected detailed human annotations for the errors in  the generated videos.

\noindent\textbf{Prompts.} We curated a dataset of 200 prompts from five sources to ensure a wide range of complexity and content. This includes 25 complex prompts from StoryEval \cite{wang2024storyeval}, 25 prompts from VBench2 \cite{zheng2025vbench2} that specifically targeting challenging categories such as ``Material," ``Mechanics," ``Human Anatomy," and ``Dynamic Spatial Relationship," and 50 from the VidProM \cite{wang2024vidprom} dataset which features creative prompts. For real-world scenes, we obtained 30 prompts from BlackSwan \cite{chinchure2025black} and 70 from LVD-2M \cite{xiong2024lvd2m}. Our prompts vary in their level of detail, and provide challenging scenarios for video generation. Prompt selection and filtering steps are in Appendix \ref{app: Data Collection:prompts}.

\noindent\textbf{Video Generation.}  We generated 600 videos from the 200 prompts using three top-performing, state-of-the-art video generation models: Veo 3 Fast\footnote{\url{https://deepmind.google/models/veo/}}, Seedance \cite{Gao2025Seedance1E}, and an open-weight model, LTX-2\footnote{\url{https://ltx.video}. Used via API, open-weights in Nov 2025.}. 

\noindent\textbf{Human Annotation.} We annotated errors using the categories defined in Table~\ref{tab:errors}. Each annotation included the error's start and end times (in seconds), a textual description, a category/type label, and a severity rating (1-5 scale) indicating how noticeable the error was. We conducted two annotation rounds. In the first round, all annotators labeled the full set of 600 videos. We then identified high-quality annotators based on two criteria: they identified at least three errors per video, and wrote descriptions of at least 8 words. In the second round, only these selected annotators re-annotated all 600 videos to ensure comprehensive, detailed error documentation. We provide details in Appendix \ref{app: Data Collection:template}.
Our annotation process was conducted through the CloudConnect by CloudResearch\footnote{\url{https://www.cloudresearch.com/}}. Each video was annotated by a qualified annotator who was compensated \$1.1 per annotation task, which sums up to \$13 per hour. 

\noindent\textbf{Data Filtering and Merging.} To reduce redundancy, we filtered and merged annotations across both rounds. We merged semantically identical errors that overlapped significantly in time (IoU $\geq$ 0.6) by using an LLM to assess semantic similarity. When merging, we retained the annotation with the more detailed description. This process yielded the final dataset.

\begin{figure}[t]
  \centering
  \includegraphics[width=1.0\linewidth]{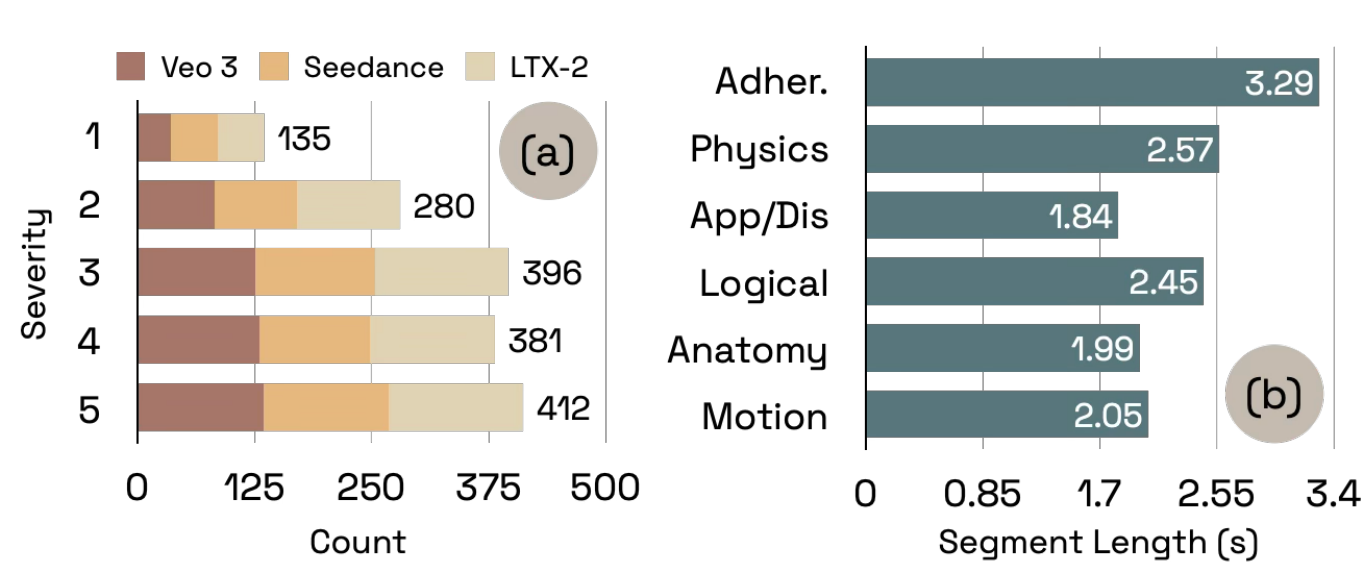}
   \caption{{\bf Analysis of T2V models on \task\ data.}}
   \label{fig:datastats}
\end{figure}

\noindent\textbf{Final Dataset Statistics.} We obtain a dataset of 1604 error annotations across 600 videos, with 484 videos having three or more errors. The annotations are detailed, with an average error reasoning length of 14.6 words. The average video length is 6.5s, while the average marked error segment is 2.49s, with a mix of local and global errors. 

\noindent\textbf{Dataset Quality.} We sample 100 error annotations from the dataset to perform human validation of the data. Human evaluators rated 93\% of error reasons and 99\% of segments as correct. Additional analysis is in Appendix \ref{app: Data Collection:qual}. 

\subsection{Analysis of Video Generators}
Our data provides us with detailed information about video generation errors from the T2V models we evaluate.

\noindent\textbf{Duration of Errors.} As illustrated in Figure~\ref{fig:datastats}, we observe that prompt adherence errors exhibit the longest duration, followed by physics and logic errors. In contrast, appearance/disappearance, motion, and body anatomy errors tend to be shorter.

\noindent\textbf{Severity of Errors.} We defined severity score as a metric of \textit{how obvious} an error is to spot. As illustrated in Figure~\ref{fig:datastats}, we find the average severity scores for Veo 3 to be 3.48, for Seedance to be 3.38, and for LTX-2 to be 3.36, which is a small difference across the models.

\noindent\textbf{Error Rates.} Veo 3 produces the fewest errors (507 annotations), followed closely by Seedance (517). The open-weight model LTX-2 produces more errors (580), highlighting a gap between commercial-only and open-weights models.

\noindent\textbf{Types.} We observe that prompt adherence errors are the most frequent, followed by physics and appearance-disappearance errors. This is in line with prior works (e.g. \cite{wang2024storyeval}), as current video generation models struggle to generate videos for complex, multi-event prompts. We share additional details in Appendix \ref{app: Data Collection:qual}.



\section{Metrics}
\label{sec:metrics}

\begin{figure}[t]
  \centering
  \includegraphics[width=1.0\linewidth]{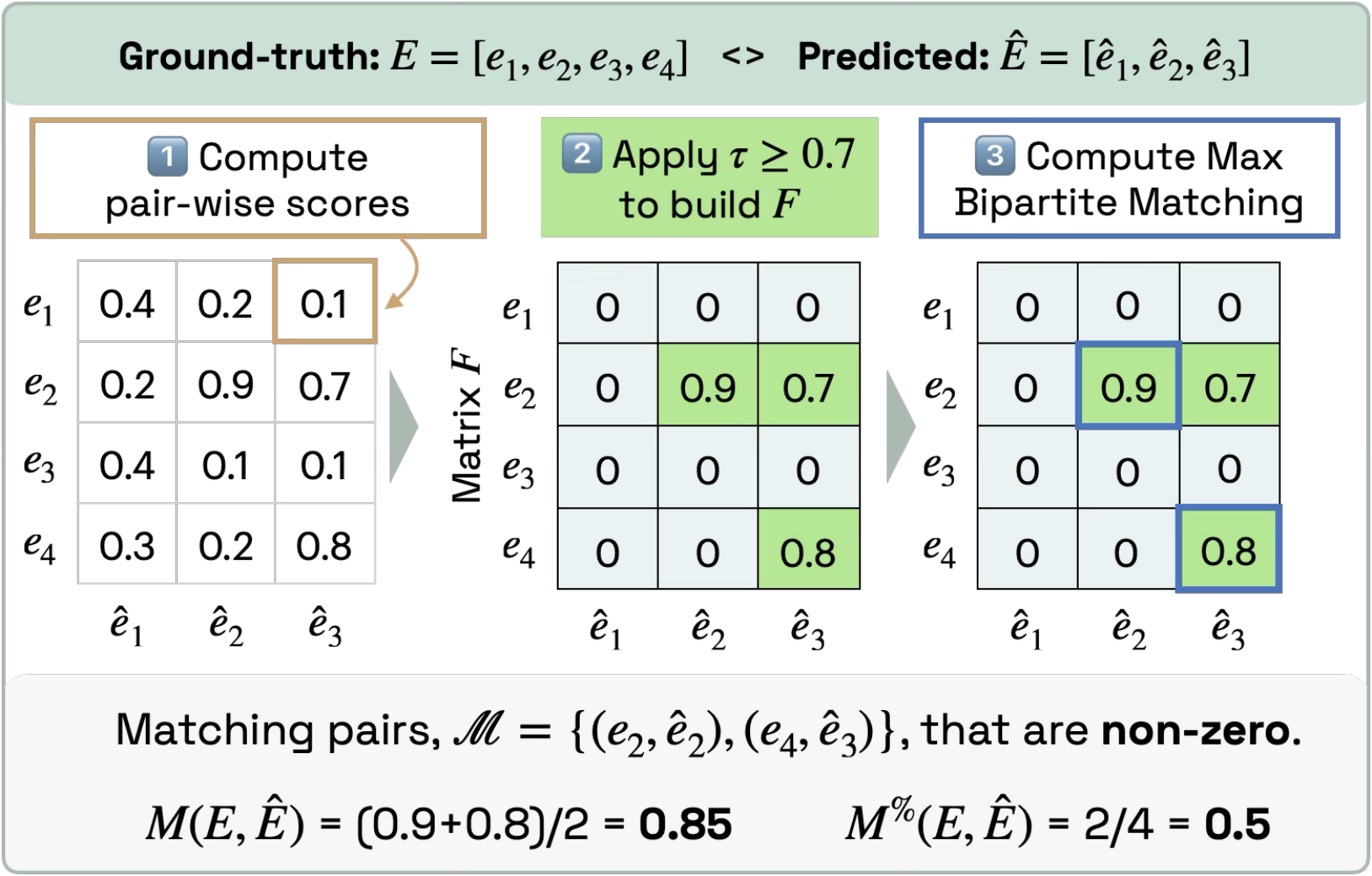}
   \caption{{\bf Example of Pairwise Matching.} After filling in matrix $F$, we compute the best match pairs. $M$ and $M^\%$ are computed subsequently.}
   \label{fig:metric}
\end{figure}

\begin{figure*}[h]
  \centering
  \includegraphics[width=1.0\linewidth]{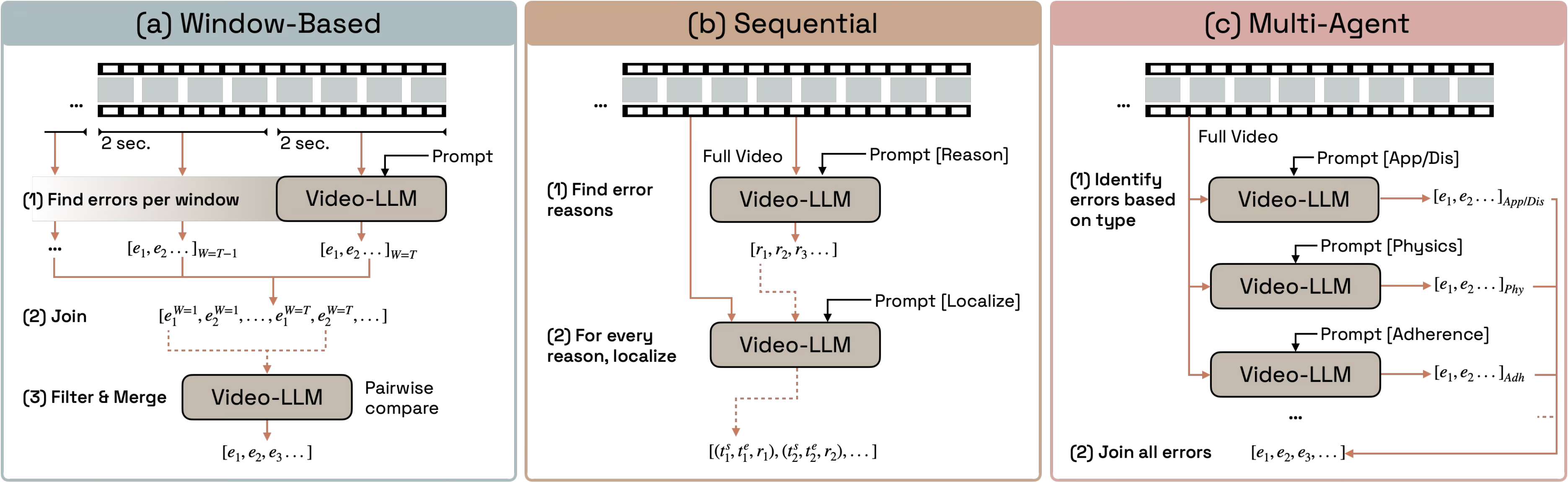}
   \caption{{\bf Our Baseline Methods.} We experiment with three approaches to perform our error reasoning and localization task.}
   \label{fig:baselines}
\end{figure*}

We define three metrics that measure how well the predicted errors $\hat{E}$ align with the human-annotated ground truth errors $E$. We then use bipartite matching to compute results across pairs of errors in the two sets, $\hat{E}$ and $E$.

\subsection{Base Metrics}

We define the following pairwise metrics to quantify alignment between each ground truth error $e_i = (t^s_i, t^e_i, c_i, r_i)$ and predicted error $\hat{e}_j = (\hat{t}^s_j, \hat{t}^e_j, \hat{c}_i, \hat{r}_j)$.

\noindent\textbf{Precision (P):} Measures the fraction of the predicted segment that overlaps with the ground truth segment:
\begin{equation}
\text{P}(e_i, \hat{e}_j) = \frac{[t^s_i, t^e_i] \cap [\hat{t}^s_j, \hat{t}^e_j]}{(\hat{t}^e_i - \hat{t}^s_i)} \in [0,1].
\end{equation}

\noindent\textbf{Recall (R):} Measures the fraction of the ground truth segment covered by the prediction: 
\begin{equation}
\text{R}(e_i, \hat{e}_j) = \frac{[t^s_i, t^e_i] \cap [\hat{t}^s_j, \hat{t}^e_j]}{(t^e_i - t^s_i)} \in [0,1].
\end{equation}
    
\noindent\textbf{Reason Similarity (S):} Measures the semantic similarity between predicted and ground-truth reasons using an LLM evaluator. We use GPT-4o-mini for this task (prompt is in Appendix \ref{app: Data Collection:prompts}). The model rates similarity from 1 to 10, representing how similar the ground truth $r_i$ is to the predicted $\hat{r}_j$,
\begin{equation}
\text{S}(e_i, \hat{e}_j) = \frac{\text{LLM-Match}(r_i, \hat{r}_j) \in [0,10]}{10} \in [0,1].
\end{equation}

When computing the metrics, we relax the need for a category label $\hat{c}_i$ match, because it is sensitive to how the error is interpreted during annotation. However, we use it to analyze differences between our baseline methods.

\subsection{Pairwise Matching Process}

To evaluate the set of predicted errors $\hat{E}$ against the ground-truth set $E$ for a given video, we must find an optimal one-to-one mapping between them. We formulate this as a maximum-weight bipartite matching problem, as shown in Fig.~\ref{fig:metric}. This process consists of three steps: (1) constructing a pairwise score matrix, (2) solving for the optimal matching, and (3) calculating the final evaluation metrics.

\noindent\textbf{Pairwise Score Matrix.}
For a given metric $\text{B} \in \{\text{P, R, S}\}$, we construct pairwise matrix, $F_{i,j}$, where each entry represents the alignment between errors $e_i$ and $\hat{e}_j$ as shown in Fig.~\ref{fig:metric}. To filter weak matches, we apply a threshold $\tau$ below which pairs are considered invalid. We build the pairwise matrix as,
\begin{equation}
F_{i,j} = \begin{cases}
      0 & \text{if } \text{B}(e_i, \hat{e}_j) < \tau\\
      \text{B}(e_i, \hat{e}_j) &\text{if } \text{B}(e_i, \hat{e}_j) \geq \tau
    \end{cases}.
\end{equation}
However, since identifying an error requires both precise temporal localization and accurate reasoning, we also define a stricter matching criterion that combines precision and semantic similarity. For our task, because the ground truth segments are generally small (avg. 2.49s), precisely marking errors is more important than recall over the marked region. We consider a prediction accurate only when both $P_{i,j} = \text{P}(e_i, \hat{e}_j)$ and $S_{i,j} = \text{S}(e_i, \hat{e}_j)$ exceed threshold $\tau$:

\begin{equation}
F_{i,j} = \begin{cases}
      0 & \text{if } P_{i,j} < \tau \vee S_{i,j} < \tau \\
      (P_{i,j}+S_{i,j})/2 &\text{otherwise}
    \end{cases}.
\end{equation}

\noindent For simplicity, we call this the S+P metric when reporting results.

\noindent\textbf{Optimal Matching.} We then solve for the maximum-weight bipartite matching (refer to Appendix. \ref{app: Metrics: Bipartite}) on $F$ to find the optimal set of matched pairs $\mathcal{M}$, only keeping non-zero pairs. This set maximizes the total alignment score $\sum_{(e_i, \hat{e}_j) \in \mathcal{M}} F_{i,j}$.



\begin{table*}[ht]
\centering
\begin{tabular}{l|c|cc|cc|cc|cc}
\toprule
 & & \multicolumn{2}{c}{\textbf{S+P}} & \multicolumn{2}{c}{\textbf{Similarity (S)}} & \multicolumn{2}{c}{\textbf{Precision (P)}} & \multicolumn{2}{c}{\textbf{Recall (R)}} \\ 
 \midrule
\textbf{@ Threshold 0.7} & \textbf{FPS} & \textbf{$M$} & \textbf{$M^\%$} & \textbf{$M$} & \textbf{$M^\%$} & \textbf{$M$} & \textbf{$M^\%$} & \textbf{$M$} & \textbf{$M^\%$} \\
\midrule
\rowcolor{gray!10} \multicolumn{10}{c}{\textbf{Zero-Shot}} \\
Qwen 3 8B-I & 2 & 0.148 & 8.0\% & 0.352 & 23.3\% & 0.472 & 27.1\% & 0.903 & 73.1\% \\
Qwen 3 30B-A3B-I & 2 & 0.137 & 6.1\% & 0.339 & 21.5\% & 0.510 & 29.9\% & \underline{0.910} & \underline{\textbf{73.9\%}} \\
Gemini 2.5 Pro & 2 & \underline{0.250} & \underline{12.6\%} & \underline{0.422} & \underline{26.7\%} & \underline{0.691} & \underline{44.3\%} & 0.836 & 68.4\% \\
\rowcolor{gray!10} \multicolumn{10}{c}{\textbf{Inference-Time Baselines}} \\
Qwen 3 8B-I - SW & 4 & 0.189 & 10.0\% & 0.386 & 24.6\% & 0.664 & 41.8\% & \underline{\textbf{0.931}} & \underline{67.6\%} \\
Qwen 3 8B-I - Seq & 2 & 0.189 & 9.3\% & 0.394 & 25.7\% & 0.645 & 39.4\% & 0.822 & 60.9\% \\
Qwen 3 8B-I - MA & 2 & \underline{0.254} & \underline{13.3\%} & \underline{0.405} & \underline{30.6\%} & \underline{0.675} & \underline{\textbf{49.8\%}} & 0.767 & 61.7\% \\
\rowcolor{gray!10} \multicolumn{10}{c}{\textbf{Human}} \\
Human & - &\textbf{0.508} & \textbf{26.8\%} & \textbf{0.599} & \textbf{34.4\%} & \textbf{0.840} & 48.5\% & 0.880 & 50.0\% \\
\bottomrule
\end{tabular}
\caption{\textbf{Comparing VLM methods on \task.} For all metrics, higher is better ($\uparrow$). We bold best performing method and underline the best method within each section. We observe that Gemini 2.5 Pro is the strongest zero-shot method, while still lagging behind humans by half. Among our inference-time strategies, MA performs the best, matching Gemini 2.5 Pro with a significantly smaller parameter model.}
\label{tab:mainres}
\end{table*}

\noindent\textbf{Final Scores.} 
From the optimal matching $\mathcal{M}$, we compute the final scores, $M(E,\hat{E})$ for video $V$ is the average over matched pairs as show in Fig.~\ref{fig:metric}:
\begin{equation}
M(E,\hat{E}) = \frac{1}{|\mathcal{M}|} \sum_{(e_i, \hat{e}_j) \in \mathcal{M}} F_{i,j}.
\label{eq:mainmetric}
\end{equation}

\noindent We also report coverage, the fraction of ground truth errors successfully matched, $M^\%$ as,
\begin{equation}
M^\%(E,\hat{E}) = |\mathcal{M}|/|E|.
\end{equation}

\section{Experimental Methods}
\label{sec:baselines}

To test the capabilites of VLMs on \task, we consider state-of-the-art VLMs capable of understanding and localizing content in videos. We evaluate Qwen3-VL \cite{qwen2025qwen3vl} (open-source) and Gemini 2.5 Pro \cite{comanici2025gemini25pushingfrontier} (closed-source). In addition to zero-shot performance on \task, we investigate a suite of inference-time approaches. Each baseline is detailed below. Prompts and additional details are in Appendix \ref{app: setup}.

\subsection{Zero-shot}
\label{sec:zero_shot}
We begin by evaluating the models’ zero-shot localization and reasoning capabilities. The model receives the full video and is prompted to produce a JSON list of detected errors. Each entry includes a timestamp interval localizing the error, the error label (Table~\ref{tab:errors}) and a short reason describing the error. Since generated videos are likely out-of-distribution (OOD) for VLMs, this setting represents a challenging evaluation of zero-shot generalization.

\subsection{Inference-Time Strategies}
\noindent\textbf{Sliding Window (SW).} To improve temporal resolution of VLMs, we extend the zero-shot baseline with a sliding-window approach as shown in Fig.~\ref{fig:baselines}(a). We partition each video into 2-second sub-clips and evaluate each sub-clip independently at 4~fps. The prompt explicitly specifies which sub-clip of the original video is being analyzed. This approach trades-off global context for improved frame density per window, allowing the model to detect more localized and transient errors. We then consolidate predictions across all sub-clips by mapping detected error segments to the global video timeline. To handle redundant detections, we use the same VLM in text-only mode to identify semantically equivalent error descriptions. Matching errors are merged by taking the union of their temporal intervals, with this process applied transitively until convergence. This gives us a unified set of error detections spanning the full video.


\noindent\textbf{Sequential (Seq): Reason then Localize.} Inspired by how humans might typically identify errors, the \textit{Reason then Localize} baseline decomposes the task into two sequential phases as shown in Fig.~\ref{fig:baselines}(b). First, a reasoning stage scans the entire video (2 fps) to identify what errors occurred and why, compiling a list without timestamps. Next, a localization stage takes each of these identified errors as a query. The model re-processes the video to find the precise temporal boundaries (start and end times) for that specific error. This design deliberately isolates the reasoning task from localization.

\noindent\textbf{Multi-Agent (MA).} This baseline adopts a divide and conquer strategy by evaluating each of the six error types independently. We query the model six times, once for each error category, using the full video in a zero-shot setting. After obtaining all six outputs, we combine them into a single list. By narrowing the focus to one error type at a time, this baseline encourages more deliberate reasoning for specific error categories.

\subsection{Experimental Setup}
\label{sec:experimental_setup}
For Qwen models, we conduct evaluations on Qwen3-VL-8B-Instruct and Qwen3-VL-30B-A3B-Instruct, using up to two NVIDIA H100 80GB GPUs. Prompts, generation configs and runtimes are in Appendix \ref{app: setup}. Gemini 2.5 Pro is accessed using the official API. We also report human performance by crowd-sourcing expert annotations for a random subset of 80 videos (see Appendix \ref{app: setup: humanbase}). 
We report results at $\tau=0.7$ for all metrics, Similarity, Precision and Recall, as well as the final score with Similarity and Precision together (S+P). For all metrics, higher is better, indicating better alignment between predictions and ground-truth annotations.


\begin{table*}[ht]
\centering
\begin{tabular}{l|cc|cc|cc|cc|cc|cc}
\toprule
\textbf{} & \multicolumn{2}{c}{\textbf{Adherence}} & \multicolumn{2}{c}{\textbf{Physics}} & \multicolumn{2}{c}{\textbf{App/Dis}} & \multicolumn{2}{c}{\textbf{Logical}} & \multicolumn{2}{c}{\textbf{Anatomy}} & \multicolumn{2}{c}{\textbf{Motion}} \\
\midrule
\textbf{} & \textbf{M} & \textbf{M\%} & \textbf{M} & \textbf{M\%} & \textbf{M} & \textbf{M\%} & \textbf{M} & \textbf{M\%} & \textbf{M} & \textbf{M\%} & \textbf{M} & \textbf{M\%} \\
\midrule
G2.5 Pro & 0.246 & 20.8\% & 0.149 & 14.4\% & 0.099 & 9.0\% & 0.083 & 5.9\% & 0.078 & 8.3\% & 0.117 & 11.1\% \\
Q3 - MA & 0.243 & 21.4\% & 0.205 & 20.3\% & 0.017 & 1.5\% & 0.132 & 11.4\% & 0.068 & 7.0\% & 0.122 & 13.0\% \\
Human & 0.532 & 45.0\% & 0.367 & 33.9\% & 0.139 & 12.1\% & 0.21 & 21.4\% & 0.153 & 17.6\% & 0.264 & 30.0\% \\
\bottomrule
\end{tabular}
\caption{\textbf{Analysis across types.} We show S+P for all. Both humans and models are stronger at detecting adherence errors over other types.}
\label{tab:types}
\end{table*}

\section{Results}
\label{sec:results}


Table~\ref{tab:mainres} summarizes the performance of different baselines.

\noindent\textbf{Zero-shot Performance.}
Among all models, \textbf{Gemini~2.5~Pro} achieves the strongest zero-shot performance, with a S+P score of 0.250 and 12.6\% of errors matched with ground truth, outperforming both Qwen-3-VL-8B-Instruct and Qwen-3-VL-30B-A3B-Instruct models by almost 2$\times$. This trend continues when only computing pairwise matching with similarity (S) and precision (P), although the improvements are narrower. It is important to address that the S+P is a more challenging but more important metric, since it requires both error reasoning and temporal localization to meet the threshold simultaneously, indicating a complete match with the ground truth annotation. Finally, we observe that recall (R) is high for all models. While this may look like a positive result, a near-perfect recall indicates a failure of localization in models like Qwen3, which often predicts the entire duration of the clip as the time segment, as seen in qualitative results (Sec. ~\ref{sec:analysis}). Comparing zero-shot performance to human performance, we observe that humans perform 2$\times$ better (0.25 versus 0.5) on the S+P metric, highlighting the clear gap between current VLMs and human-level understanding.

\noindent\textbf{Inference-time strategies.}
We observe that all inference-time strategies improve performance over zero-shot Qwen3. Notably, the multi-agent (MA) approach is very effective, closely matching (S and P) or surpassing (S+P) the zero-shot performance of the large-scale Gemini 2.5 Pro on \task. This multi-agent approach achieves the highest performance in precision (49.8\%) and a modest decrease in recall compared to the base model. The two-step sequential approach proves to be the least successful. This shows that identifying and localizing errors are closely related, and disentangling them may prove challenging for models. Finally, we observe that the window-based approach achieves very high recall scores. This could be an artifact of merging errors across windows through semantic similarity matching of errors. 

\noindent\textbf{Human.} We observe that, while humans perform \task\ relatively better than models, they still do not achieve very high scores $M$ or $M^\%$ against ground-truth error annotations. This is likely because the \task\ dataset contains error annotations from multiple rounds, combining the efforts of multiple human annotators in identifying and localizing errors. As each human approaches this task with a slightly different lens due to cognitive biases, it is likely that they will identify different errors. Additionally, no single human annotator would write a large number of errors, as it is tedious. These reasons explain the low scores. However, these low scores do not reflect negatively on the dataset quality; the quality of our annotations is independently measured and reported in Sec. \ref{sec:dataset}. 

\section{Analysis}
\label{sec:analysis}
\begin{figure}[t]
  \centering
  \includegraphics[width=1.0\linewidth]{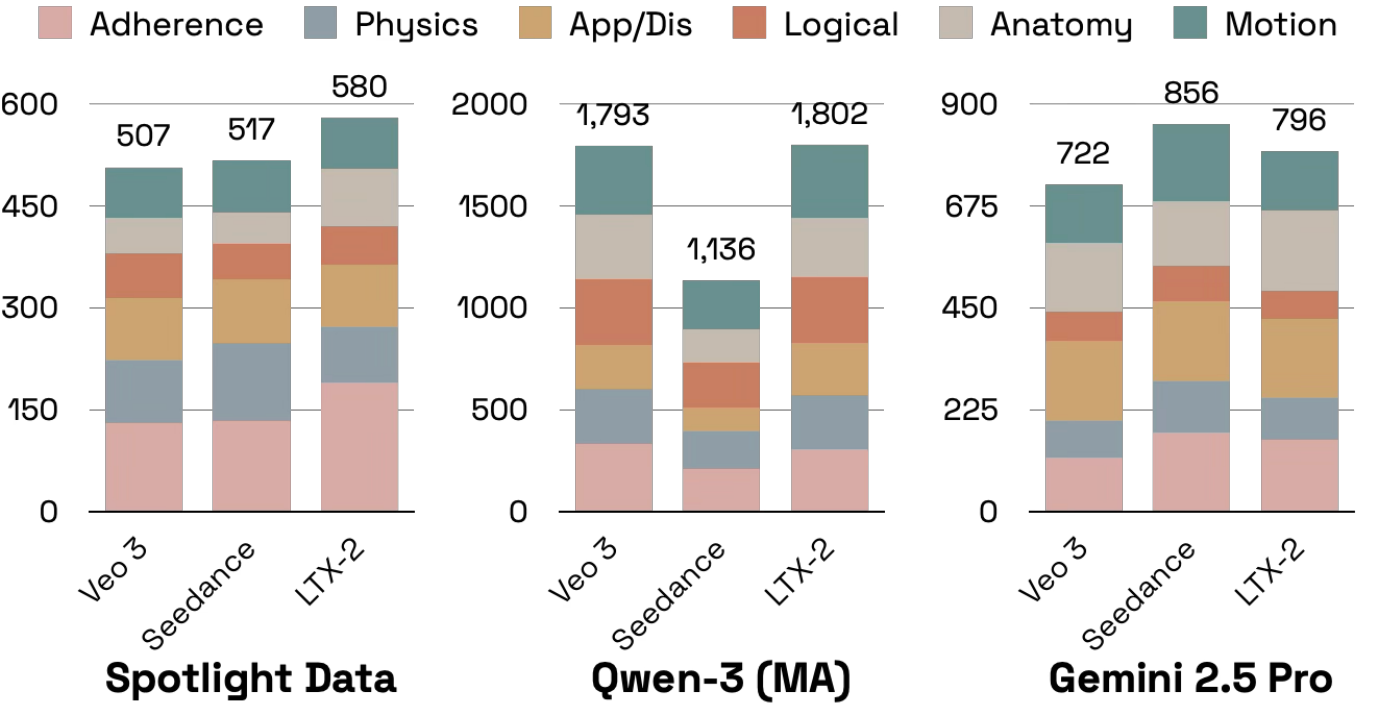}
   \caption{{\bf Comparing \# of errors detected by humans and models.} Comparing against ground truth data in \task, we observe that both models produce different error distributions for the three T2V generators.}
   \label{fig:error_analysis}
\end{figure}

\begin{figure}[t]
  \centering
  \includegraphics[width=1.0\linewidth]{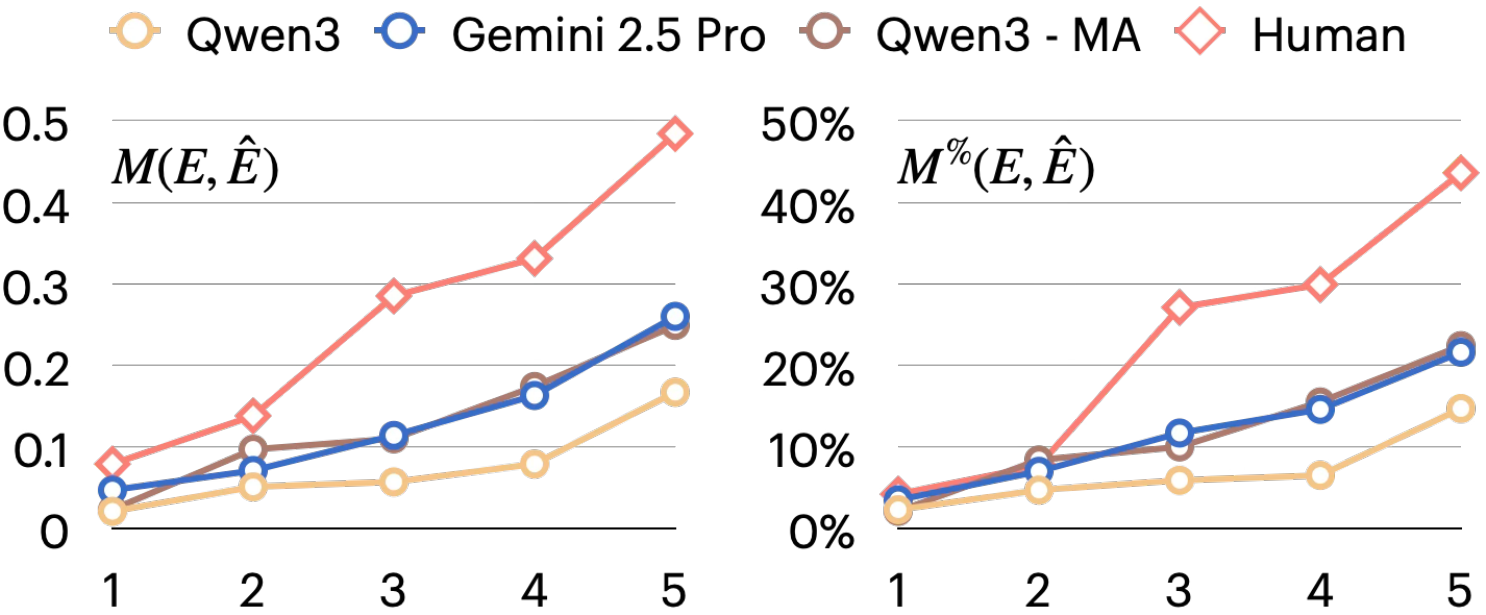}
   \caption{{\bf Severity of error vs. S+P.} We observe that more severe errors are more easily detected by all methods.}
   \label{fig:severitychart}
\end{figure}

\begin{figure*}[t!]
  \centering
  \includegraphics[width=1.0\linewidth]{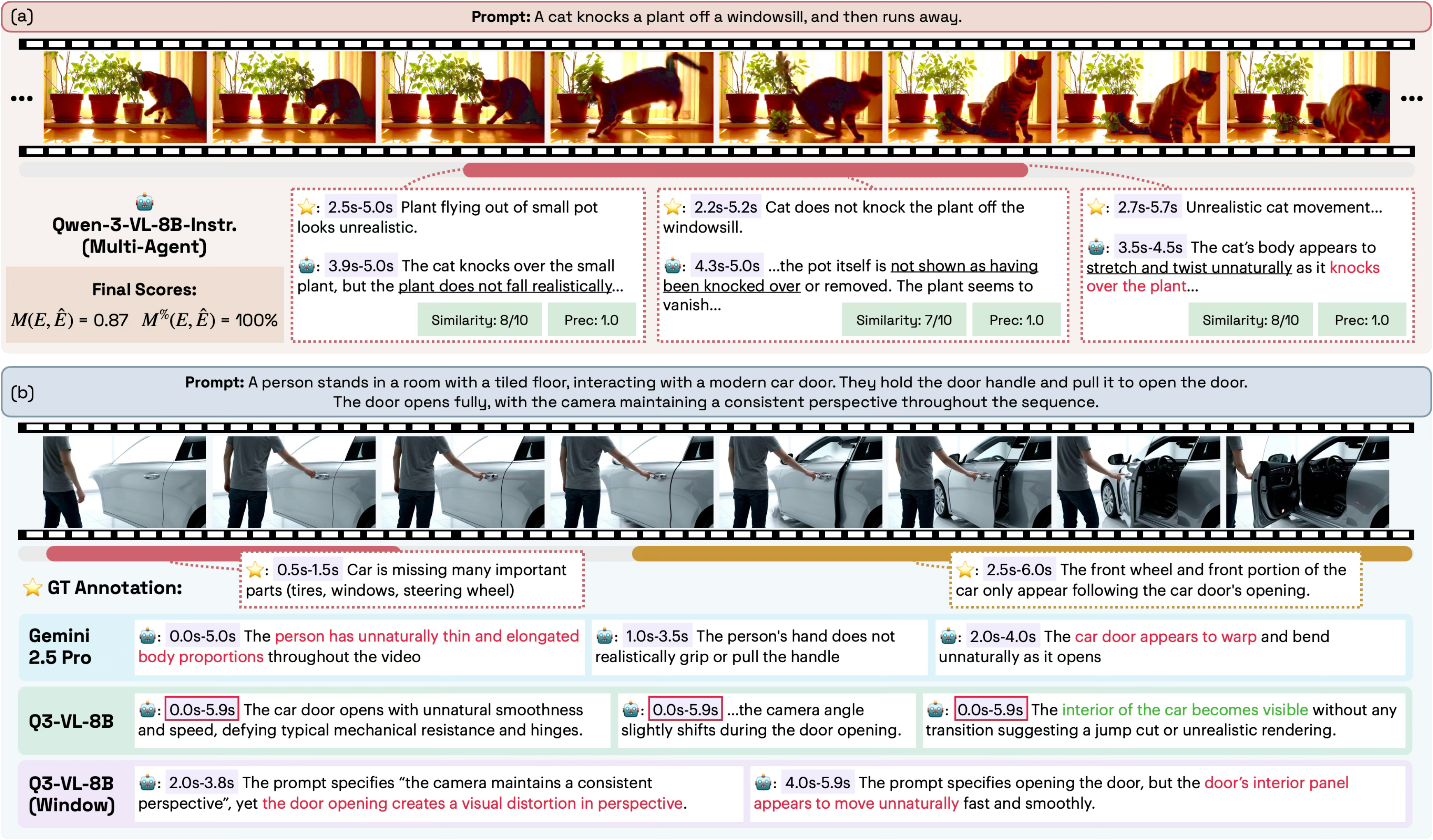}
   \caption{{\bf Qualitative Examples.} In (a), we show a video where Qwen3-MA is able to detect all ground truth errors, achieving $M^\% = 100\%$. In (b), we show ways in which models fail at \task\, through incorrect error reasoning and imprecise localization.  Not all predicted errors are shown due to space constraints.}
   \label{fig:qual}
\end{figure*}


\subsection{Analysis on Errors}

\noindent\textbf{Which video generators have more errors according to humans vs. VLMs?}
Fig.~\ref{fig:error_analysis}  compares error detection patterns between humans, and the best performing open-sourced (Qwen3-MA) and closed-source (Gemini 2.5 Pro) models. We observe notable differences in which generators each evaluator finds most problematic. Humans  find LTX-2 videos to be the most error-prone, with ~600 of errors. In contrast, models show different preferences: Qwen3-MA identifies Seedance as the least error-prone generator, detecting 1,136 errors compared to about 1,800 errors for Veo 3 and LTX-2. Gemini 2.5 Pro finds Veo 3 to be the least error-prone, with 722 errors, although the relative skew across T2V models is smaller.

\noindent\textbf{How does error severity influence accuracy?}
As shown in Fig.~\ref{fig:severitychart}, we observe that more severe and obvious errors are easier to detect for all methods. Qwen3-MA and Gemini 2.5 Pro show similar improvements across severity levels, while the base Qwen3 lags behind. However, humans show more substantial improvements on severity levels 3-5 compared to all methods.

\noindent\textbf{What types of errors do humans vs. VLMs detect?}
The charts in Fig.~\ref{fig:error_analysis} include type-specific splits across humans and models. Humans detect more adherence errors (about 28\%) vs. Qwen3-MA (18\%) and Gemini (19\%). Gemini predicts more appearance/disappearance errors (22\% vs. 17\% for humans) while Qwen predicts motion errors (20\% vs. 14\% for humans). These differences indicate how different models may perceive errors differently.

\noindent\textbf{How does error type influence detection accuracy?}
Table \ref{tab:types} presents a performance breakdown across six error types. We observe that both models and humans can more reliably detect errors related to prompt adherence over other types. Qwen3-MA is stronger at detecting physics and logical errors (0.2 and 0.13) than Gemini (0.14 and 0.08), whereas Gemini is stronger at detecting appearance and disappearance errors (0.099 vs 0.017). 



\subsection{Qualitative Examples}
\label{Examples:Qual}

Figure \ref{fig:qual} presents qualitative samples of model predictions. (a) shows Qwen3-MA achieving perfect matching $M^\% = 100\%$, with predicted errors exhibiting high similarity scores to ground truth error reasons and with precise temporal localization. Task decomposition enables more comprehensive multi-error detection. However, our final score $M$ accurately captures that the predicted reasons are not an exact match, with Qwen3-MA predicting conflicting statements about the plant being knocked over (marked in red).

However, (b) reveals the limitations of current VLMs through a failure case where all models miss the primary error in a relatively simple scene. The ground truth annotation identifies that a large portion of the car is invisible throughout most of the video, only appearing after the door opens. Gemini 2.5 Pro incorrectly focuses on the person's body (which appears normal), while Qwen3 fails at temporal localization entirely, marking the full video in each error. The window-based approach narrows the error segments in this case, but fails to identify any errors correctly. 
\section{Conclusion}
\label{sec:conclusion}
\task\ is a novel task to identify and localize fine-grained errors in AI-generated videos. Through comprehensive human annotation of 1,604 errors across 600 videos from three state-of-the-art T2V generators, we reveal where and when modern T2V models fail. We then evaluate whether current VLMs can diagnose the same errors, uncovering key limitations such as precise temporal reasoning for detecting subtle and localized inconsistencies. Our experiments show humans outperform state-of-the-art models by 2×, and while inference-time strategies can improve performance, a substantial gap remains. Addressing these limitations is crucial for building robust evaluation tools for video generation systems.

While we have collected high-quality error annotations over all videos, we acknowledge that our dataset may not cover all possible errors in these videos due to the differences in how humans perceive these videos, along with cognitive biases. Still, \task\ remains a challenging benchmark for VLMs, driving progress towards VLMs with stronger temporal understanding and diagnostic capabilities. We hope to enable stronger, fine-grained and human-aligned video generation evaluation.

{
    \small
    \bibliographystyle{ieeenat_fullname}
    \bibliography{main}
}

\clearpage
\setcounter{page}{1}
\maketitlesupplementary
\renewcommand{\thesection}{\Alph{section}}
\setcounter{section}{0}


\section{Data Collection: Additional Information}
\label{app: Data Collection}
\subsection{Examples from prompt sources}
\label{app: Data Collection:prompts}

We use 5 sources for the 200 prompts. 

\noindent\textbf{(25/200) StoryEval:} This dataset contains prompts with multiple parts to a story. Since this is a curated prompt set, we do not apply any filtering, and simply sample randomly. Some examples are:
\begin{itemize}
    \item \texttt{A kangaroo joey peeks out from its mother's pouch, hops out, and then runs away.}
    \item \texttt{A cat climbs a tree, and then lounges on a branch.}
\end{itemize}

\noindent\textbf{(25/200) VBench2:} VBench uses a suite of prompts to standardize evaluation across different T2V models. We sample 25 prompts from the Material, Mechanics, Human Anatomy, and Dynamic Spatial Relationship sets. Some examples:
\begin{itemize}
    \item \texttt{A man is doing yoga.} (Human Anatomy)
    \item \texttt{A squirrel is on the left of a rock, then the squirrel jumps to the top of the rock.} (Dynamic Spatial Relationship)
\end{itemize}

\noindent\textbf{(50/200) VidProM:} VidProM is a large-scale set of 1.67M unique T2V prompts from real users. This set includes creative prompts. We first filter prompts based on whether they may contain unsafe or NSFW material (based on the classifications already provided in the dataset\footnote{\url{https://huggingface.co/datasets/WenhaoWang/VidProM}}, and then sample 50 prompts from the ShareVeo3\footnote{\url{https://huggingface.co/datasets/WenhaoWang/ShareVeo3}} subset containing 6763 prompts. Some examples:
\begin{itemize}
    \item \texttt{a stunning house in the middle of a lush green forest with rivers flowing around.}
    \item \texttt{at a fork in the road, horse avoids the path that the sign indicates leads to the finish line and takes another route.}
\end{itemize}

\noindent\textbf{(30/200) BlackSwan:} BlackSwanSuite contains descriptions surprising events from short real-world videos. We use the descriptions in the Reporter split of the Validation set, and filter away descriptions that are shorter than 10 words. We then sample 30 prompts. Some examples:
\begin{itemize}
    \item \texttt{The guy jumps in the air with his skateboard, but lands unevenly on the skateboard, causing him to lose balance with the skateboard, and the guy slips off and onto the ground while the skateboard rolls away.}
    \item \texttt{The tree flies out of the man's hands and he then falls to the ground. Someone asks if he is okay as he starts to get up.}
\end{itemize}

\noindent\textbf{(70/200) LVD-2M:} This dataset contains 2 million long videos (around 10s) and is annotated with temporally dense captions. We only sample prompts from the \texttt{data/ytb\_600k\_720p.csv} split provided on GitHub\footnote{\url{https://github.com/SilentView/LVD-2M}} and use the rewritten captions provided in the dataset. Examples include:
\begin{itemize}
    \item \texttt{A hippopotamus walks towards a body of water, then submerges itself, and finally emerges and moves away from the water's edge. The surrounding natural landscape features trees and shrubs.}
    \item \texttt{A person's hands manipulate a deck of playing cards on a surface, spreading them out and shuffling or sorting them.}
\end{itemize}

\subsection{Human Annotation}
\label{app: Data Collection:template}

\paragraph{Video Generation.} For video generation, we use the \texttt{fal.ai} API to access Veo 3 Fast (\url{fal-ai/veo3/fast}), LTX-2 Fast (\url{fal-ai/ltxv-2/text-to-video/fast}), and Seedance 1.0 Pro (\url{fal-ai/bytedance/seedance/v1/pro/text-to-video}). We always use default settings, except that we randomly vary video lengths for Veo 3 and LTX-2 by setting length to 6 seconds or 8 seconds.

\begin{figure*}[t]
  \centering
  \includegraphics[width=0.7\linewidth]{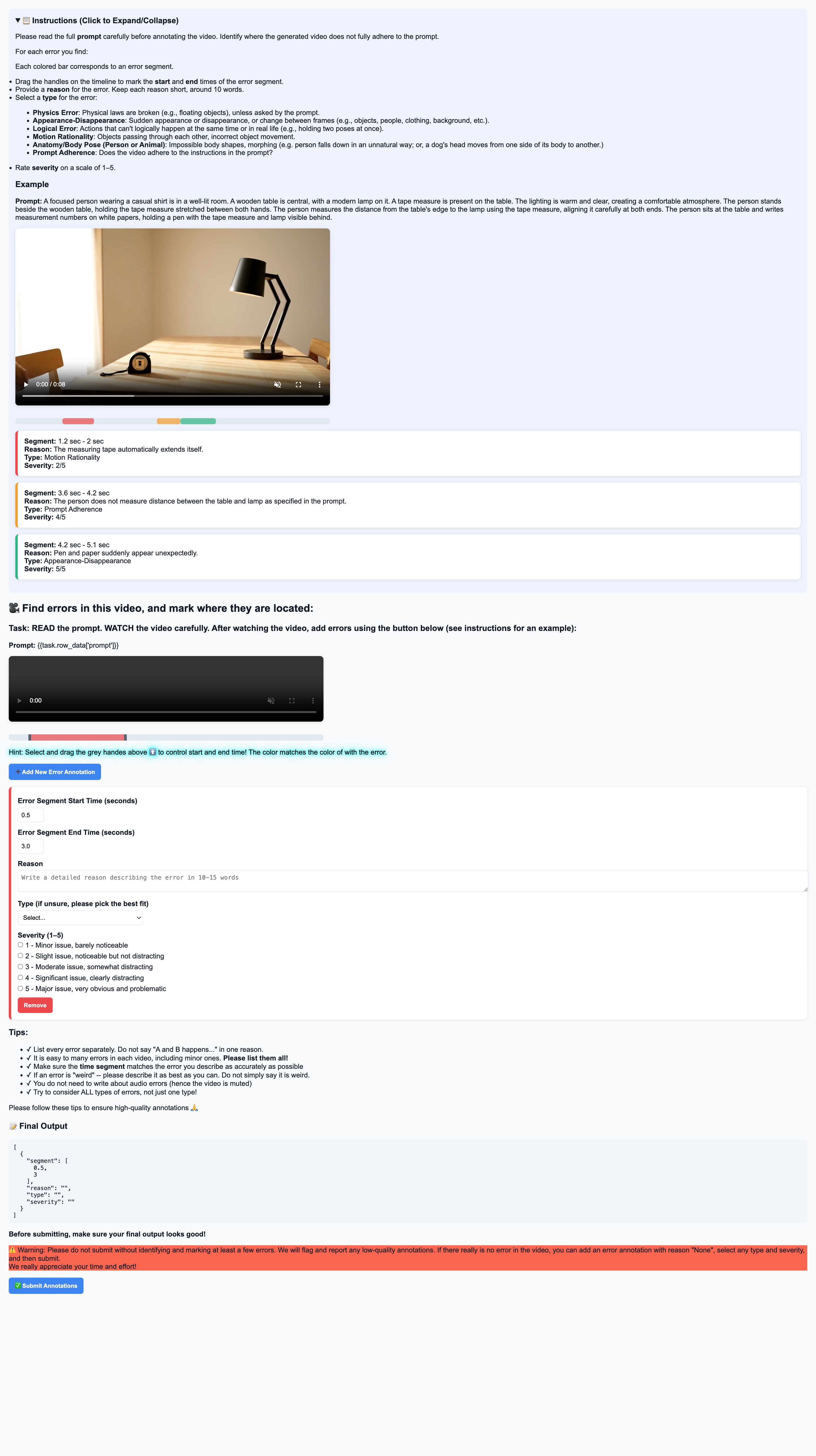}
   \caption{{\bf Main Template.} This is the template we use for collecting annotations for \task.}
   \label{fig:template:main}
\end{figure*}

\paragraph{Template for Annotation.} The template for human annotation is in Fig. \ref{fig:template:main}.

\begin{figure}[ht]
  \centering
  \includegraphics[width=1.0\linewidth]{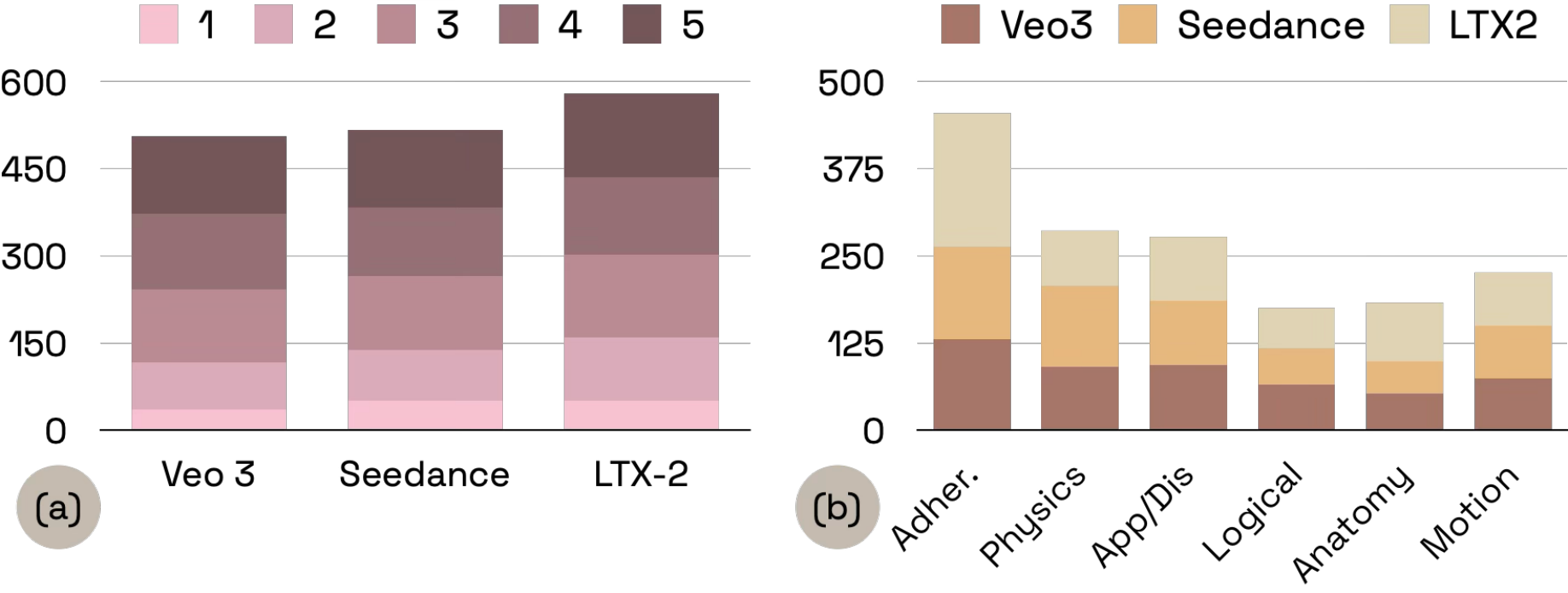}
   \caption{{\bf \task\ Analysis.} (a) shows severity across different models. (b) shows types of errors.}
   \label{fig:app:datainfo}
\end{figure}

\subsection{Analysis}
\label{app: Data Collection:qual}

In Fig. \ref{fig:app:datainfo}, we show (a) how severity scores are distributed across models and (b) which types of errors are found in the generated videos. We observe that semantic prompt adherence errors are the most common, followed by physics and appearance-disappearance.

\begin{figure*}[t]
  \centering
  \includegraphics[width=1.0\linewidth]{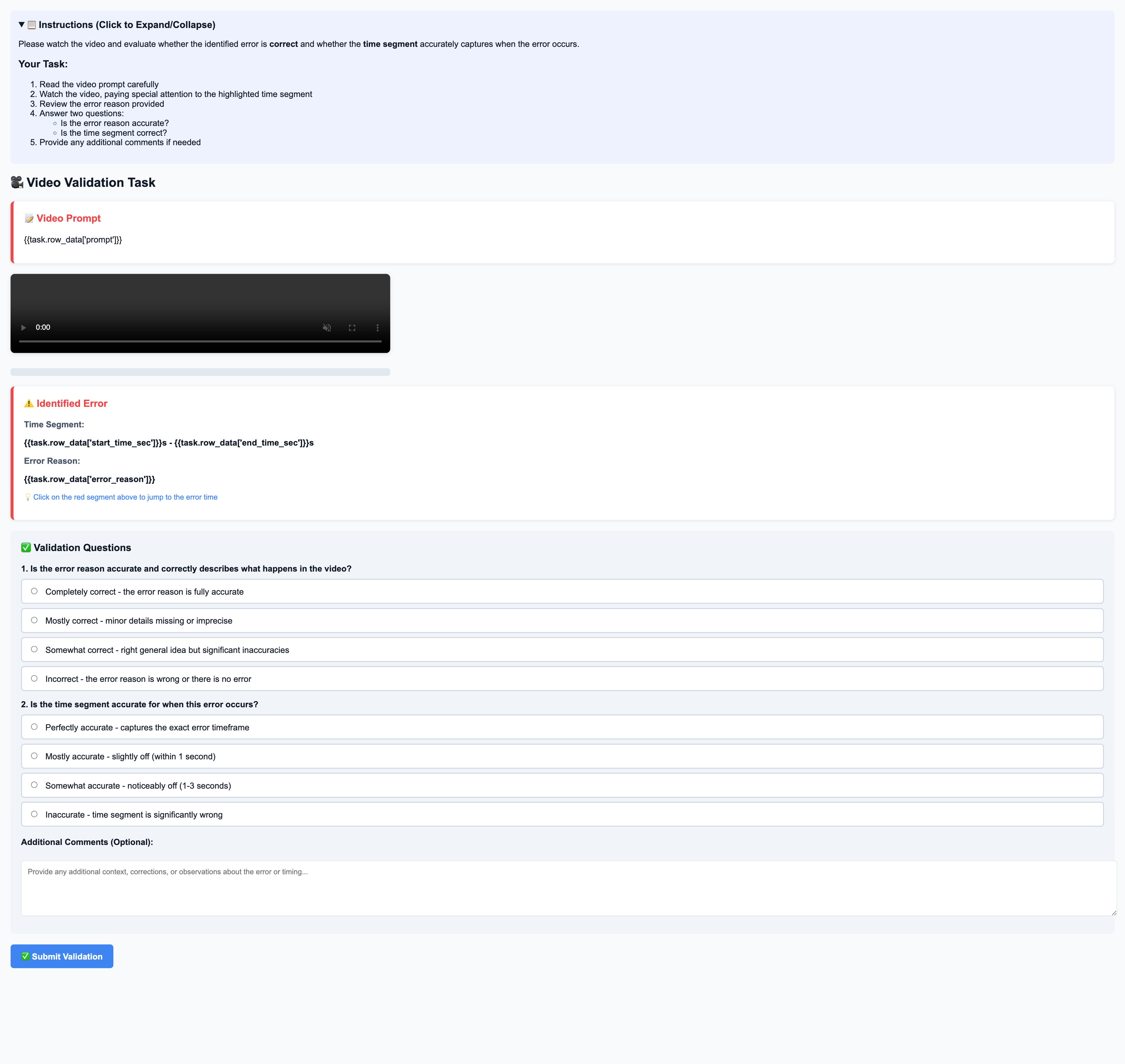}
   \caption{{\bf Human Eval Template.} To conduct human evaluation on the quality of the dataset, we ask users to rate each error independently.}
   \label{fig:template:heval}
\end{figure*}

\paragraph{Data Quality.} We try to ensure that our data is of high quality. Nonetheless, some errors may be present. We use the template in Fig. \ref{fig:template:heval} to rate the quality of each error we sampled for this analysis. Annotators rated 93\% of error reasons and 99\% of segments are marked as \textit{at least} ``somewhat correct" or ``somewhat accurate", with 86\% and 93\% being \textit{at least} ``mostly correct" and ``mostly accurate" respectively. Incorrect errors stemmed from annotators misrepresenting the nature or severity of the error in their reasoning, or video quality issues (like aberrations or blurring which is hard to visually interpret).

\section{Metrics: Additional Info}
\label{app: Metrics}

\subsection{Reason Similarity}
\label{app: Metrics: LLM Match}

The prompt we use with GPT-4o-mini for Reason Similarity is:

\begin{prompt}
You are a semantic similarity evaluator. You compare two error statements, both of which are describing errors about the same video generated by AI. \\
Rate how similar in meaning the two error statements are, ignoring differences in wording, phrasing, or length. Focus only on whether they describe the same situation, cause, or event. \\

Use the 0-10 scale, and return any integer from 1 to 10. Here's a rough explanation of the scale: \\
0 = The errors are different, low score \\
5 = The errors overlap in meaning or implications, but key details are missing, medium score \\
10 = The two errors have essentially identical meanings (even if wording or phrasing differs), high score \\

Example 1: \\
Error A: ``At no point does the dog trip and roll forward, as requested by the prompt." \\
Error B: ``The dog does not trip or roll forward across the sand; it runs continuously." \\
Output: 9 \\

Example 2: \\
Error A: ``The helmet/head of the motorcyclist collapse in the mud in an unnatural way." \\
Error B: ``The rider and motorcycle disappear completely into the mud without any visible transition or realistic impact motion." \\
Output: 6 \\

Example 3: \\
Error A: ``yellow shrimp's leg movements aren't natural, they appear/disappear unnaturally and sometimes overlap" \\
Error B: ``The cleaner shrimp is not actively cleaning the fish; it is merely perched on its head." \\
Output: 2 \\

Task: \\
Error A: [reason1] \\
Error B: [reason2] \\
Output only a single integer from 0 to 10 with no explanation or additional text.
\end{prompt}

\begin{figure*}[t]
  \centering
  \includegraphics[width=1.0\linewidth]{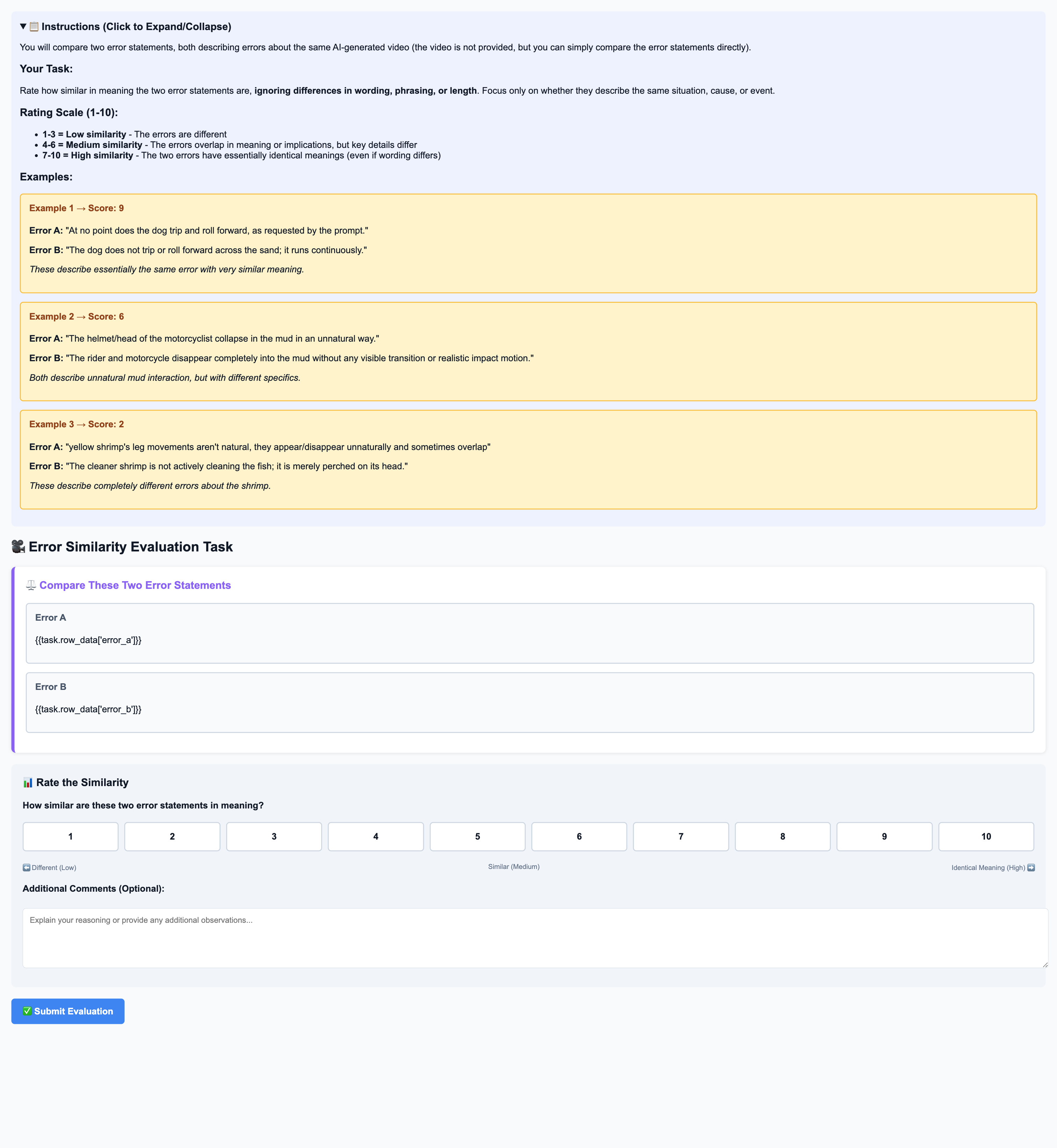}
   \caption{{\bf Reason Similarity Eval Template.} We evaluate the LLM-Match metric using this template, based on the Reason Similarity prompt.}
   \label{fig:template:llmmatch}
\end{figure*}

To evaluate how reliably our metric works, we do human evaluation. The template for human evaluation follows the above prompt and is provided in Fig \ref{fig:template:llmmatch}. To perform human evaluation, we first select 100 pairs. Because most text does not match, there is a much higher likelihood that any randomly sampled pair (ground truth, model generated) of reason text has a low reason similarity score. Therefore, we use LLM-provided scores to first obtain pairs in each rating score bucket from 1-10. We then sample 10 pairs from each bucket, giving us 100 pairs. We use all pairs across all models shown in Table \ref{tab:mainres} to ensure diversity. We obtain a Spearman correlation of \textbf{+0.77} with humans, indicating \textbf{strong correlation}.

\subsection{Bipartite Match}
\label{app: Metrics: Bipartite}

We use SciPy's Linear Sum Assignment (\url{https://docs.scipy.org/doc/scipy/reference/generated/scipy.optimize.linear_sum_assignment.html}) to do bipartite matching on the matrix. Since this function, by default, tries to minimize weight, we instead apply a negative sign to all values in matrix $F$ to compute the maximum weight matching instead.  

\subsection{Other Metrics}
\label{app: Metrics: Other}

\paragraph{IoU:} Instead of precision and recall, we considered using IoU for temporal segment matching. However, we found that IoU scores do not correctly reflect the issue of full-video segments being generated (very high recall, as shown in Tab. \ref{tab:mainres}) with some models like Qwen3. We therefore report precision and recall separately. 

\paragraph{Type-classification metric:} Because of the nature of error types being up to the interpretation of the annotator or model, we do not consider error types as a part of our metrics, although it is a part of the prompt in \ref{app: setup: prompts}.

\section{Experimental Setup: Additional Info}
\label{app: setup}
\subsection{Prompts}
\label{app: setup: prompts}

\paragraph{Zero-shot.} Here's the prompt for zero-shot:
\begin{prompt}
You are watching a video generated by an AI model using this prompt: [prompt] \\

Your task is to describe and localize all visual generation errors in this video. \\

Focus on the following error types: \\
1. physics\_violation: Physical laws are broken (e.g., floating objects), unless asked by the prompt. \\
2. object\_appearance\_disappearance: Sudden appearance or disappearance, or change between frames (e.g., objects, people, clothing, background, etc.). \\
3. logical\_error: Actions that can't logically happen at the same time or in real life (e.g., holding two poses at once). \\
4. motion\_rationality: Objects passing through each other, incorrect object movement. \\
5. anatomy\_body: Impossible body shapes (Person or Animal), morphing (e.g. person falls down in an unnatural way; or, a dog's head moves from one side of its body to another.) \\
6. semantic\_prompt\_adherence: Does the video adhere to the instructions in the prompt or not \\

Return a JSON list of all errors found, by specifying the time segments (in seconds) for localization, the reason, and type for each error. Time segments can vary in length and can overlap if needed. All error reasons must be unique. Keep the reasons short but clear. \\
Example Output: \\
\texttt{[ \\
  {"Segment": "1.2 sec - 2 sec", "Reason": "The measuring tape automatically extends itself.", "Type": "motion\_rationality"}, \\
  {"Segment": "1.0 sec - 4.9 sec", "Reason": "The person does not measure the distance between the table edge and the lamp correctly.", "Type": "semantic\_prompt\_adherence"}, \\
  {"Segment": "4.3 sec - 5.1 sec", "Reason": "Objects appear out of thin air.", "Type": "object\_appearance\_disappearance"}, \\
  {"Segment": "0.0 sec - 5.9 sec", "Reason": "The background does not match the scene described in the prompt.", "Type": "semantic\_prompt\_adherence"} \\
... more errors ... \\
]}
\end{prompt}

\paragraph{Sliding Window.} Here's the prompt for sliding window. For each window, we mention start and end time for each segment, and total length of the video:
\begin{prompt}
You are watching a part of a video generated by an AI model. This part is from [S] sec to [E] sec of a [L] sec video. The video is generated using this prompt: [prompt] \\

Your task is to describe and localize all visual generation errors in this video. \\

Focus on the following error types: \\
1. physics\_violation: Physical laws are broken (e.g., floating objects), unless asked by the prompt. \\
2. object\_appearance\_disappearance: Sudden appearance or disappearance, or change between frames (e.g., objects, people, clothing, background, etc.). \\
3. logical\_error: Actions that can't logically happen at the same time or in real life (e.g., holding two poses at once). \\
4. motion\_rationality: Objects passing through each other, incorrect object movement. \\
5. anatomy\_body: Impossible body shapes (Person or Animal), morphing (e.g. person falls down in an unnatural way; or, a dog's head moves from one side of its body to another.) \\
6. semantic\_prompt\_adherence: Does the video adhere to the instructions in the prompt or not \\

Return a JSON list of all errors found, by specifying the time segments (in seconds) for localization, the reason, and type for each error. Time segments can vary in length and can overlap if needed. All error reasons must be unique. Keep the reasons short but clear. \\
Example Output: \\
\texttt{[ \\
  {"Segment": "1.2 sec - 2.0 sec", "Reason": "The measuring tape automatically extends itself.", "Type": "motion\_rationality"}, \\
  {"Segment": "1.0 sec - 1.9 sec", "Reason": "The person does not measure the distance between the table edge and the lamp correctly.", "Type": "semantic\_prompt\_adherence"}, \\
  {"Segment": "0.3 sec - 1.1 sec", "Reason": "Objects appear out of thin air.", "Type": "object\_appearance\_disappearance"}, \\
  {"Segment": "0.0 sec - 2.0 sec", "Reason": "The background does not match the scene described in the prompt.", "Type": "semantic\_prompt\_adherence"} \\
... more errors ... \\
]}
\end{prompt}

Then, to combine errors across windows by using the same model, and asking it whether two error statements (from two different videos) are the same: 
\begin{prompt}
You are a semantic similarity evaluator. You are given two error reasons for AI generated videos. Do these``error reason" statements describe the same error in the video? Ignore differences in wording, phrasing, or length. \\

Reason A: [reason1]\\
Reason B: [reason2]\\

Output only 0 for false and 1 for true.
\end{prompt}

If two errors are the same, then we can merge their time segments, giving us errors that span across multiple windows.

\paragraph{Sequential.} First, we only as the model to reason about the errors, without doing localization:
\begin{prompt}
You are watching a video generated by an AI model using this prompt: [prompt] \\

Your task is to describe all visual generation errors in this video. \\

Focus on the following error types: \\
1. physics\_violation: Physical laws are broken (e.g., floating objects), unless asked by the prompt. \\
2. object\_appearance\_disappearance: Sudden appearance or disappearance, or change between frames (e.g., objects, people, clothing, background, etc.). \\
3. logical\_error: Actions that can't logically happen at the same time or in real life (e.g., holding two poses at once). \\
4. motion\_rationality: Objects passing through each other, incorrect object movement. \\
5. anatomy\_body: Impossible body shapes (Person or Animal), morphing (e.g. person falls down in an unnatural way; or, a dog's head moves from one side of its body to another.) \\
6. semantic\_prompt\_adherence: Does the video adhere to the instructions in the prompt or not \\

Return a JSON list of all errors found, by specifying the reason and type for each error. All error reasons must be unique. Keep the reasons short but clear. \\
Example Output: \\
\texttt{[ \\
  {"Reason": "The measuring tape automatically extends itself.", "Type": "motion\_rationality"}, \\
  {"Reason": "The person does not measure the distance between the table edge and the lamp correctly.", "Type": "semantic\_prompt\_adherence"}, \\
  {"Reason": "Objects appear out of thin air.", "Type": "object\_appearance\_disappearance"}, \\
  {"Reason": "The background does not match the scene described in the prompt.", "Type": "semantic\_prompt\_adherence"} \\
... more errors ... \\
]}
\end{prompt}

We then do localization, for every error that is detected:
\begin{prompt}
You are given a video and a description of a specific visual generation error.\\

Here is the error: [reason] \\

Your task is to precisely localize the time segment in the video where the aforementioned error occurs. \\

Guidelines: \\
- Focus only on the given error description. \\
- You should specify start and end times in seconds, by looking at the frames and considering when this error could have occurred in the video. \\
- Be as precise as possible with your start and end times. \\

Return your output as a time segment string in the following format. Do not include any additional text or explanation.\\
Example Output: \texttt{2.8 sec - 3.2 sec}
\end{prompt}

\paragraph{Multi-Agent.} We have a different prompt for each error type. To reduce repetition, we only show the template for the prompt:
\begin{prompt}
You are watching a video generated by an AI model using this prompt: [prompt] \\

Your task is to describe all visual generation errors related to [error-type], and localize them in the video. \\

To identify [error-type] errors, look for: [error-desc]. \\

Return a JSON list of all errors found, specifying the time segments (in seconds) for localization, the reason for each error.
If there are no [error-type] errors in the video, return an empty list. \\

Example Output: \\
\texttt{[ \\
  {"Segment": "1.2 sec - 2 sec", "Reason": "..."} \\
]}
\end{prompt}
Here, the [error-type] and [error-desc] are replaced with the error types and descriptions in our error taxonomy (Tab. \ref{tab:errors}, and similar to what is present in the zero-shot prompt).

\begin{figure*}[t]
  \centering
  \includegraphics[width=1.0\linewidth]{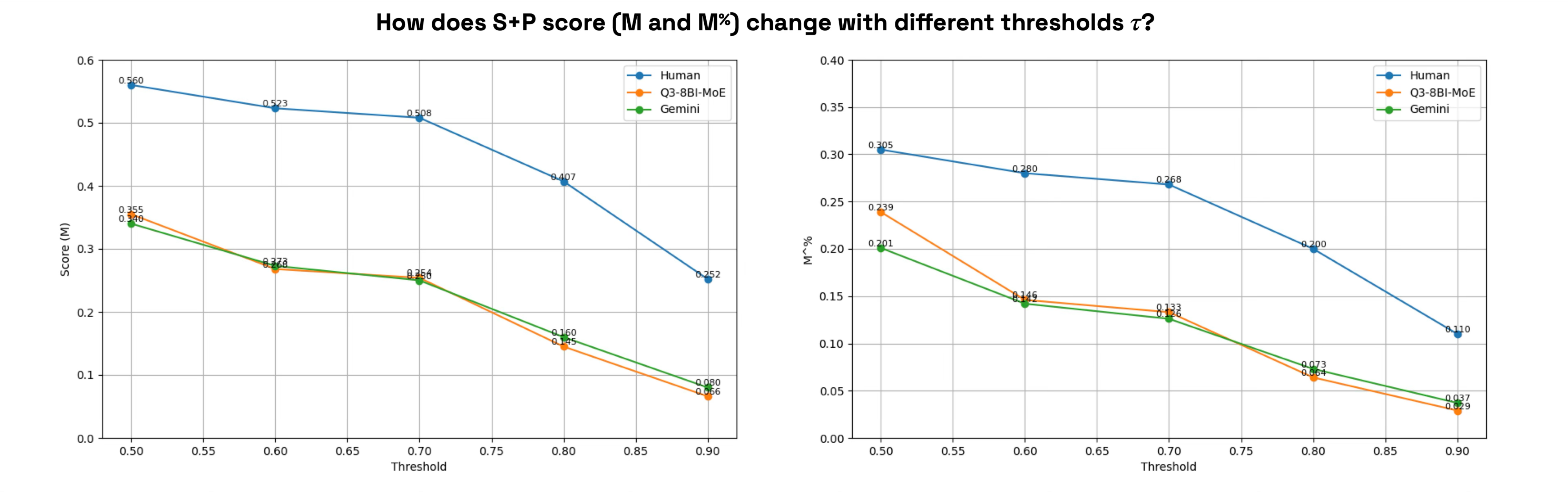}
   \caption{{\bf Hyperparameter: Threshold.} We show S+P results on three models while varying threshold $\tau$, and observe that at smaller thresholds, more of the pairs are considered valid (hence the higher scores), but the gap between model and human performance remains. With very high thresholds, the matched pairs are few, leading to a drop in both model and human performance.}
   \label{fig:app:thresh}
\end{figure*}

\begin{figure*}[t]
  \centering
  \includegraphics[width=1.0\linewidth]{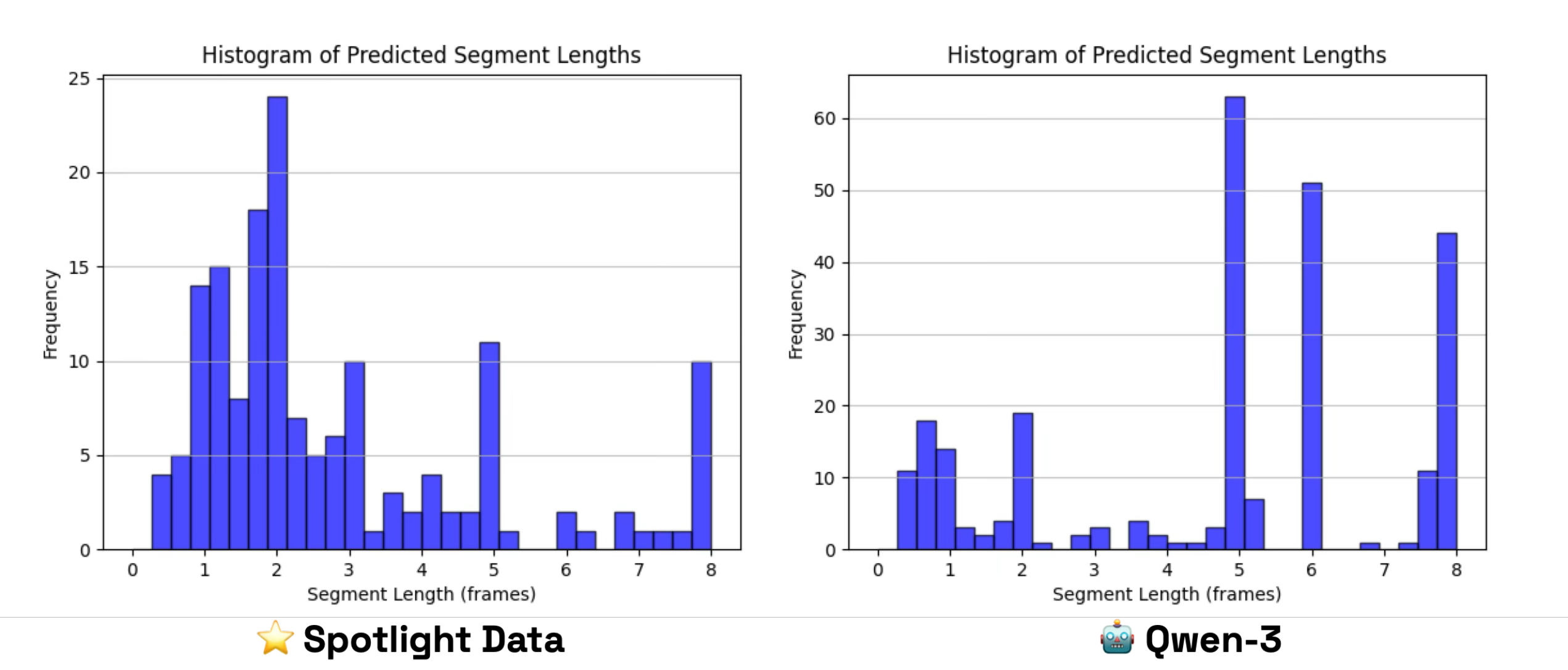}
   \caption{{\bf Segment Lengths.} In Table \ref{tab:mainres}, we observe very high recall scores for models. This is because models predict the full length of the video very often (as shown with the spikes at 5s, 6s, and 8s, which are the three lenghts of our videos).}
   \label{fig:app:seglens}
\end{figure*}

\subsection{Setup}
\label{app: setup: algo}

\paragraph{Qwen 3 Family.} For Qwen 3 models, we use Hugging Face setup and configurations for both 8B\footnote{\url{https://huggingface.co/Qwen/Qwen3-VL-8B-Instruct}} and 30B-A3B\footnote{\url{https://huggingface.co/Qwen/Qwen3-VL-30B-A3B-Instruct}} models. We directly input the video using local file paths in the conversation and set the \texttt{fps} setting in the processor based on which of our inference-time methods we are using.
We run all Qwen 3 models on up to two Nvidia H100 GPUs. On average, to do inference on the full \task\ dataset, it takes 1.5 hrs. 

\paragraph{Gemini.} For the Gemini 2.5 Pro model, we use the default setup provided in the Gemini API documentation. As we only conduct zero-shot experiments, we set \texttt{fps} to 2.

\subsection{Analysis}
\label{app: setup: runtimes}

\paragraph{Why $\tau = 0.7$?} We analyze models and human performance on the task across different values of $\tau$ in Fig. \ref{fig:app:thresh}. For smaller thresholds, although scores are higher because of more valid matches, the gap between models and humans remains. With very high thresholds, the matched pairs are few, leading to a drop in both model and human performance. Therefore, we choose 0.7 as a reasonable middle-ground.

\paragraph{Segment Lengths.} We plot the segment lengths predicted by humans and by models in Fig \ref{fig:app:seglens}. As stated in Sec. \ref{sec:results}, we observe that models fail to localize errors entirely, by more frequently predicting the full duration of the clip as the segment. This is indicated by the spikes at 5s, 6s, and 8s, which are the lengths of the videos in the dataset.

\subsection{Human Expert Baseline}
\label{app: setup: humanbase}

For the human baseline, we re-annotate 80 randomly selected videos across all three video generators with different high quality annotators (filtered based on length of reason annotation ($>8$ words) and number of errors ($\geq3$), as described in Sec \ref{sec:dataset}) who did not have a chance to participate in round 2 of the data annotation process. This ensured that while quality is maintained, we collect unique annotations.

\end{document}